\documentclass[conf]{new-aiaa}
\usepackage[utf8]{inputenc}

\usepackage{graphicx}
\usepackage{amsmath}
\usepackage[version=4]{mhchem}
\usepackage{siunitx}
\usepackage{longtable,tabularx}
\usepackage{cleveref}
\crefname{assumption}{assumption}{assumptions}
\usepackage{caption}
\usepackage{subcaption}
\usepackage{multirow}
\usepackage{booktabs}
\allowdisplaybreaks
\usepackage[makeroom]{cancel}
\usepackage{fancyhdr}

\fancypagestyle{firstpage}{
    \fancyhf{}
    \fancyfoot[C]{%
        \scriptsize Copyright \copyright\ 2026 by the American Institute of Aeronautics and Astronautics, Inc. All rights reserved.\\
        Published in AIAA SciTech 2026: https://doi.org/10.2514/6.2026-1959
    }

}

\newtheorem{assumption}{Assumption}
\DeclareMathOperator{\diag}{diag}

\newcommand{\R}{\mathbb{R}}
\newcommand{\N}{\mathbb{N}}
\usepackage{acro}
\DeclareAcronym{fie}{
    short = FIE,
    long = Full Information Estimation,
}
\DeclareAcronym{mhe}{
    short = MHE,
    long = Moving Horizon Estimation,
}
\DeclareAcronym{kf}{
    short = KF,
    long = Kalman Filter,
}
\DeclareAcronym{ekf}{
    short = EKF,
    long = Extended Kalman Filter,
}
\DeclareAcronym{rgas}{
    short = RGAS,
    long = robustly globally asymptotically stable,
}
\DeclareAcronym{ras}{
    short = RAS,
    long = robustly asymptotically stable,
}
\DeclareAcronym{gas}{
    short = GAS,
    long = globally asymptotically stable,
}
\DeclareAcronym{qp}{
    short = QP,
    long = Quadratic Program
}
\DeclareAcronym{lti}{
    short = LTI,
    long = linear time-invariant
}
\DeclareAcronym{ltv}{
    short = LTV,
    long = linear time-varying
}
\DeclareAcronym{pcm}{
    short = PCM,
    long = Prediction-Correction Method
}
\DeclareAcronym{pcip}{
    short = PCIP,
    long = Prediction-Correction Interior-Point
}
\DeclareAcronym{l1ao}{
    short = $\mathcal{L}_1$-AO,
    long = $\mathcal{L}_1$ Adaptive Optimizer
}
\DeclareAcronym{rmse}{
    short = RMSE,
    long = root mean squared error
}
\title{Moving Horizon Estimation for Quadrotors:\\
An \acl{l1ao} Approach}

\author{
    Thinh Q. Nguyen\footnote{Corresponding Author, Master's Student, Department of Mechanical Science and Engineering, The Grainger College of Engineering.},
    Minkyung Kim\footnote{Doctoral Student, Department of Mechanical Science and Engineering, The Grainger College of Engineering.},
    Sandeep Banik\footnote{Postdoctoral Research Associate, Department of Mechanical Science and Engineering, The Grainger College of Engineering.},
    Jinrae Kim\footnote{Postdoctoral Research Associate, Department of Mechanical Science and Engineering, The Grainger College of Engineering.},
    and
    Naira Hovakimyan\footnote{Professor, Department of Mechanical Science and Engineering, The Grainger College of Engineering.}
}
\affil{University of Illinois Urbana-Champaign, Urbana, IL 61801}

\begin{document}

\maketitle

\begin{abstract}
\textit{Abstract}---\acl{mhe} (\acs{mhe}) is a state estimation method based on finite-horizon optimization that can offer higher accuracy at the cost of increased computation compared to Kalman filter-based approaches. We present a linear smoothing \acs{mhe} formulation as a dense \acl{qp} (\acs{qp}), and a solver consisting of a continuous-time Newton's method augmented with the \acl{l1ao} (\acs{l1ao}). While \acs{mhe} is inherently time-varying, conventional approaches treat it as a sequence of independent, time-invariant problems and employ iterative solvers at each time step, which can be both inaccurate and computationally burdensome. In contrast, time-varying solvers track the optimal solution with fewer iterations by exploiting the temporal evolution of the problem, thereby reducing the computational load. In this research, we enhance both the performance and efficiency of \acs{mhe} through a time-varying solver with an \ac{l1ao} augmentation that compensates for the prediction inaccuracy, which is common in practice due to noisy sensors and the lack of prior knowledge of the system. Simulation results on a quadrotor platform show that the \acs{l1ao}-augmented approach solves the \acs{mhe} optimization problem more efficiently than the baseline time-invariant solver and achieves higher estimation accuracy under challenging conditions, compared with both the \acl{ekf} and the standard \acs{mhe}.
\end{abstract}

\section{Nomenclature}

{\renewcommand\arraystretch{1.0}
\noindent\begin{longtable*}{@{}l @{\quad=\quad} l@{}}
$n$ & number of system states \\
$m$ & number of control inputs \\
$p$ & number of system outputs (measurements) \\
$Q, R, P_0$ & estimator weighting matrices \\
$V_f$ & \acs{fie} objective function \\
$V$ & \acs{mhe} objective function \\
$N$ & \acs{mhe} horizon length \\
$x, y$ & system states and outputs \\
$\chi, \eta$ & decision variables corresponding to states and outputs \\
$\hat{x}, \hat{y}$ & \emph{optimal} decision variables corresponding to states and outputs \\
$w, v$ & process and measurement noises \\
$\omega, \nu$ & decision variables corresponding to process and measurement noises \\
$\hat{\omega}, \hat{\nu}$ & \emph{optimal} decision variables corresponding to process and measurement noises \\
$\mathbf{z}$ & decision variable of the \acs{mhe} \acs{qp} \\
$\mathbf{H}, \mathbf{f}$ & symmetric matrix and vector of the \acs{mhe} \acs{qp} objective function \\
$\mathbf{A}_s$ & a user-selected diagonal Hurwitz matrix \\
$T_s$ & sampling period \\
$\omega_c$ & low-pass filter cutoff frequency
\end{longtable*}}

\setcounter{table}{0}
\thispagestyle{firstpage}

% -----------------------------------

\section{Introduction}\label{sec:introduction}

% ------ Quadrotors ------
Quadrotors have emerged as highly versatile platforms for a wide range of applications, from aerial surveying to search-and-rescue missions. Their performance critically depends on the underlying feedback control policy, which requires an \emph{accurate} estimate of the system state. However, in practice, state estimation for quadrotors is inherently challenging due to model inaccuracies and noisy measurements from both proprioceptive (e.g., IMUs) and exteroceptive (e.g., cameras) sensors. Moreover, limited onboard computational resources introduce additional challenges for real-time estimation. 

% ------ EKF ------
\ac{kf}–based approaches are commonly adopted due to their computational efficiency. For unconstrained, linear dynamical systems subject to Gaussian noise, the \ac{kf} is a statistically optimal estimator. For nonlinear systems, a common estimation approach is to linearize the system dynamics and apply the linear \ac{kf} framework, which is known as the \acl{ekf} (\acs{ekf}). Owing to its computational efficiency and ease of implementation, the \ac{ekf} is widely used and has demonstrated good performance on quadrotors \cite{munguia2019ekf, bohm2021combined}. However, the \ac{ekf} has several significant limitations, including the inability to explicitly incorporate state constraints into the problem and the potential for inconsistent covariance propagation due to functional linearization.

% ------ MHE ------
\acl{mhe} (\acs{mhe}) \cite{rawlings2017_MPCbook_4} presents a promising alternative by leveraging finite-horizon optimization to improve estimation accuracy. \ac{mhe} is emerging as a practical approach for online state estimation in nonlinear systems \cite{kuhl2011real, pfeiffer2021computationally, kang2024fast}. Conventional \ac{mhe} implementations solve a new finite-horizon optimization problem at each time step using iterative numerical solvers. However, on small platforms with limited hardware, such as quadrotors, solving a separate optimization problem at each time step can be computationally burdensome. Instead, we formulate \ac{mhe} as a time-varying optimization problem, i.e., with an objective function and constraints that depend explicitly on time. Unlike standard receding-horizon formulations that repeatedly solve time-invariant problems, time-varying optimization methods exploit temporal continuity to track the optimal solution over time, reducing computation time without requiring full convergence at each time step.

A widely adopted method for time-varying optimization problems is the \ac{pcm} \cite{simonetto2016class, baumann2004newton, pcip2018}. The key idea of \ac{pcm} is to decompose the tracking of the time-varying optimal solution into a prediction step and a correction step. The prediction step propagates the current optimizer forward in time using the known temporal variation of the objective function and constraints. In contrast, the correction step refines this prediction by enforcing optimality and feasibility at the current time. By alternating between these two steps, \ac{pcm} enables continuous tracking of the optimizer trajectory as the problem evolves, making it well-suited for real-time implementation. However, the performance of \ac{pcm} relies on the accuracy of its predictions, which may not hold in real-world settings due to imperfect prior knowledge.

% ------ L1AO ------
To overcome these limitations, we leverage the \acl{l1ao} (\acs{l1ao}) \cite{kim2025L}. \ac{l1ao} augments a baseline time-varying optimization solver with an adaptive update law and low-pass filtering, enabling fast and robust tracking of time-varying optimal solutions. In the context of \ac{mhe}, we present a time-varying approach with an \ac{l1ao} augmentation to solve this estimator optimization problem, obtaining a state estimator that is more robust and less computationally demanding than standard \ac{mhe} with time-invariant iterative solvers such as OSQP \cite{osqp}, making it well suited for real-time state estimation on quadrotors.

The remainder of this paper is organized as follows. \Cref{sec:preliminaries} presents the preliminaries on \ac{mhe} formulations and quadrotor dynamics. \Cref{sec:methodology} details the proposed methodology, and \Cref{sec:simulation} presents simulation studies demonstrating the effectiveness of the proposed approach. \Cref{sec:conclusion} summarizes the main findings and discusses future directions.

% -----------------------------------

\section{Preliminaries}\label{sec:preliminaries}

\subsection{\acl{mhe}}\label{sec:preliminaries_mhe}
Consider the discrete-time nonlinear dynamical system
\begin{equation}\label{eq:dynamics_system_variables}
    x^+ = f(x,u,w), \qquad y=h(x)+v,
\end{equation}
where $x\in\R^n$ is the system state, $u\in\R^m$ is the control input, $y\in\R^p$ is the output (measurement), $w\in\R^n$ is the process noise, and $v\in\R^p$ is the measurement noise. To distinguish between the true system and the state estimates, let
\begin{equation}\label{eq:dynamics_decision_variables}
    \chi^+ = f(\chi,u,\omega), \qquad y=h(\chi)+\nu,
\end{equation}
where $(\chi,\omega,\eta,\nu)$ are the decision variables in the estimator optimization problem corresponding to the system variables $(x,w,y,v)$. The optimal decision variables are denoted $(\hat{x},\hat{w},\hat{y},\hat{v})$. Given an initial guess $\overline{x}_0$ of the true initial state $x_0$, we first consider the \acl{fie} (\acs{fie}) problem at time $T$

\begin{equation}\label{eq:fie_problem}
\begin{aligned}
    \min_{\chi_{0},\{\omega_k\}_{k=0}^{T-1}} &V_{f}\big(\chi_{0}, \{\omega_k\}_{k=0}^{T-1}, T\big) \\
    \text{s.t. } \eqref{eq:dynamics_decision_variables},
\end{aligned}
\end{equation}
where $\chi_0$ is the estimate of $x_0$, $\{\omega_k\}_{k=0}^{T-1}:=(\omega_0^\top, \ldots, \omega_{T-1}^\top)^\top$ denotes the sequence of estimates of the true process noise $\{w_k\}_{k=0}^{T-1}$, and
\begin{equation}\label{eq:fie_objective_generic}
    V_{f}\big(\chi_{0}, \{\omega_k\}_{k=0}^{T-1}, T\big) = \ell_x(\chi_0-\overline{x}_0) + \sum_{i=0}^{T-1}\ell_\omega(\omega_i) + \sum_{i=0}^{T}\ell_\nu(\nu_i),
\end{equation}
where $\ell_x(\cdot)$ is the prior cost, $\ell_\omega(\cdot)$ and $\ell_\nu(\cdot)$ are the stage costs that penalize the process and measurement noises, respectively. For unconstrained linear time-invariant systems subject to zero-mean normally distributed process and measurement noises, the \ac{fie} problem \eqref{eq:fie_problem} with quadratic prior and stage costs is equivalent to the \ac{kf} \cite{rawlings2017_MPCbook_4}. The major problem with \ac{fie} is that as time $T$ progresses, this optimization problem grows in size and quickly becomes intractable. Instead, we consider a moving-horizon approach, where optimization is performed only over a finite number of decision variables. For $T<N$, we can simply solve the \ac{fie} problem. For $T\geq N$, the \acl{mhe} problem with quadratic costs is
\begin{equation}\label{eq:mhe_problem}
\begin{aligned}
    \min_{\chi_{T-N},\{\omega_k\}_{k=T-N}^{T-1}} &V\big(\chi_{T-N}, \{\omega_k\}_{k=T-N}^{T-1}, T\big) \\
    \text{s.t. } \eqref{eq:dynamics_decision_variables},
\end{aligned}
\end{equation}where
\begin{equation}\label{eq:mhe_objective_general}
    V\big(\chi_{T-N}, \{\omega_k\}_{k=T-N}^{T-1}, T\big)
    = \frac{1}{2} \big\|\chi_{T-N}-\hat{x}_{T-N}\big\|_{P_{T-N}^{-1}}^2
    + \frac{1}{2}\sum_{i=T-N}^{T-1} \big\|\omega_i\big\|_{Q^{-1}}^2 
    + \frac{1}{2}\sum_{i=T-N}^{T} \big\|\nu_i\big\|_{R^{-1}}^2,
\end{equation}
and $P_{T-N}\in\R^{n \times n}$, $Q\in\R^{n \times n}$, and $R\in\R^{p \times p}$ are weighting matrices that represent the covariances of $\chi_{T-N}$, $\{\omega_k\}_{k=T-N}^{T-1}$, and $\{\nu_k\}_{k=T-N}^{T}$, respectively.

The first term in \eqref{eq:mhe_objective_general} is known as the \emph{arrival cost}, which penalizes the difference between $\chi_{T-N}$, the estimate of $x_{T-N}$, and $\hat{x}_{T-N}$, the previously obtained optimal estimate. To estimate the arrival cost, \cite{UNGARALA2009_mhe_arrival_cost} presented several methods, including the use of \ac{ekf} and sampling-based filters such as the Unscented Kalman Filter and the Particle Filter. It has been demonstrated that sampling-based methods, which avoid functional linearization, can offer superior accuracy in the computation of the arrival cost parameters.

Based on how the arrival cost parameters are defined, we can classify \ac{mhe} into two types: \emph{filtering} and \emph{smoothing} \cite{rawlings2017_MPCbook_4}. In the \emph{filtering} \ac{mhe}, we define $\hat{x}_{T-N}:=\hat{x}_{T-N|T-N}$ and $P_{T-N}:=P_{T-N|T-N-1}$, i.e., using the optimal estimate and covariance obtained exactly at their respective point in time. On the other hand, the \emph{smoothing} \ac{mhe} uses the \emph{latest} optimal estimate and covariance that are obtained at time $T-1$, i.e. $\hat{x}_{T-N}:=\hat{x}_{T-N|T-1}$ and $P_{T-N}:=P_{T-N|T-1}$. The latter can be calculated using the backward iteration given in \cite{rawlings2017_MPCbook_4,rao2000,rao2001}
\begin{equation}\label{eq:mhe_smoothing_P_backward_iteration}
\begin{gathered}
    P_{k|T} = P_{k|k} + M_k\Big(P_{k+1|T} - P_{k+1|k}\Big)M_k^\top, \\
    M_k = P_{k|k}A_k^\top P_{k+1|k}^{-1}.
\end{gathered}
\end{equation}
Another factor we need to consider when using the smoothing \ac{mhe} is that $\hat{x}_{T-N|T-1}$ is the solution of the \ac{mhe} problem at time $T-1$, which accounts for the measurement sequence $\{y_k\}_{y=T-N-1}^{T-1}$, while in the current \ac{mhe} problem \eqref{eq:mhe_problem}, we are accounting for $\{y_k\}_{k=T-N}^{T}$, so there is an overlap of the data $\{y_k\}_{k=T-N}^{T-1}$. Simply solving \eqref{eq:mhe_problem} with the objective \eqref{eq:mhe_objective_general} would cause oscillations and undesirable steady-state errors. In the case of linear time-invariant systems, \cite{rawlings2017_MPCbook_4,rao2000,rao2001} provide an adjustment to account for the overlapping data $\{y_k\}_{k=T-N}^{T-1}$ by subtracting a term, quadratic in $\chi_{T-N}$, from the objective \eqref{eq:mhe_objective_general}. Although this method is seemingly straightforward, some complications arise during implementation. For linear systems, \eqref{eq:mhe_problem} can be written as a standard \acl{qp} (\acs{qp}) and solved very efficiently using existing convex optimization solvers, possibly through parsers. However, some parsers, such as CVXPY \cite{diamond2016cvxpy}, do not support objective functions with a subtraction of quadratic terms, which they treat as non-convex. In \Cref{sec:methodology_mhe_qp}, we will derive the expressions for the smoothing \ac{mhe} as a standard \ac{qp}, which is supported by most standard convex optimization solvers and parsers.

One benefit of the smoothing \ac{mhe} is that it eliminates the undesirable periodic spikes in the state estimates given a poor initial guess $\overline{x}_0$, a behavior observed in the filtering \ac{mhe}. This is due to the fact that the smoothing \ac{mhe} uses the \emph{latest} optimal estimate for its arrival cost, so the effect of the poor initial guess is reduced after each time step, instead of being repeated every $N$ time steps as in the filtering \ac{mhe}. In addition, compared to the filtering \ac{mhe}, the smoothing \ac{mhe} also provides faster error convergence given a poor initial guess.

% -----------------------------------

\subsection{Quadrotor dynamics}\label{sec:preliminaries_quadrotor}
The equations of motion for a quadrotor are given by
\begin{subequations}\label{eq:quadrotor_dynamics}
\begin{align}
    \dot{p} &= q, \\
    \dot{q} &= \frac{T}{m}R(\Theta)e_3 - ge_3, \\
    \dot{\Theta} &= W(\Theta)\Omega, \\
    \dot{\Omega} &= J^{-1}\big(\tau - \Omega\times(J\Omega)\big),
\end{align}
\end{subequations}
where $p\in\R^3$ and $q\in\R^3$ are the position and velocity of the quadrotor's center of mass in world frame, respectively, $m$ is the quadrotor's mass, $T$ is the collective thrust, $e_3=(0,0,1)^\top$, $g$ is the gravitational acceleration, $\Theta=(\phi,\theta,\psi)^\top\in\R^3$ is the Euler angles in world frame, $\Omega\in\R^3$ is the angular velocity in body frame, $\tau\in\R^3$ is the moment in body frame, $J\in\R^{3\times3}$ is the moment of inertia of the quadrotor, and
\begin{equation*}
    R(\Theta) = \begin{pmatrix}
        c_\psi c_\theta &
        c_\psi s_\theta s_\phi - s_\psi c_\phi &
        c_\psi s_\theta c_\phi + s_\psi s_\phi \\[4pt]
        s_\psi c_\theta &
        s_\psi s_\theta s_\phi + c_\psi c_\phi &
        s_\psi s_\theta c_\phi - c_\psi s_\phi \\[4pt]
        -s_\theta &
        c_\theta s_\phi &
        c_\theta c_\phi
    \end{pmatrix}, \qquad
    W(\Theta) = \begin{pmatrix}
        1 & s_\phi t_\theta & c_\phi t_\theta \\
        0 & c_\phi & -s_\phi \\
        0 & \frac{s_\phi}{c_\theta} & \frac{c_\phi}{c_\theta}
    \end{pmatrix},
\end{equation*}
where $s_\phi$ denotes $\sin(\phi)$, $c_\theta$ denotes $\cos(\theta)$, and $t_\theta$ denotes $\tan(\theta)$. Let the state vector be $x=(p^\top, q^\top, \Theta^\top, \Omega^\top)^\top$, and suppose noisy measurements of the position $p$ and angular velocity $\Omega$ are available, for instance, provided by a GPS sensor and a gyroscope. The measurement model is then given by
\begin{equation}
    y = Cx + v,
\end{equation}
where
\begin{equation*}
    C=\begin{pmatrix}
        \mathbf{I}_3 & \mathbf{0}_{3 \times 6} & \mathbf{0}_{3 \times 3} \\
        \mathbf{0}_{3 \times 3} & \mathbf{0}_{3 \times 6} & \mathbf{I}_3
    \end{pmatrix}, \qquad 
    v=\begin{pmatrix}
        v_{gps} \\ v_{gyro}
    \end{pmatrix},
\end{equation*}
and $v_{gps}\in\R^3$ and $v_{gyro}\in\R^3$ are the measurement noise of the GPS sensor and gyroscope, respectively.

% -----------------------------------

\section{Methodology}\label{sec:methodology}

\subsection{Linear smoothing \ac{mhe} as \acl{qp}}\label{sec:methodology_mhe_qp}

To formulate a linear \ac{mhe} problem, we approximate the nonlinear system \eqref{eq:dynamics_decision_variables} as a linear time-varying one by linearization around some trajectory. A good candidate is $\{\hat{x}_{k|T-1}\}_{k=T-N}^{T-1}$, which is obtained from the solution of the \ac{mhe} problem at $T-1$, $(\hat{x}_{T-N-1|T-1}^\top, \{\hat{\omega}_{k|T-1}^\top\}_{k=T-N-1}^{T-2})^\top$. Linearization around this trajectory of previous optimal estimates produces the time-varying linear system

\begin{equation}\label{eq:dynamics_LTV}
\begin{aligned}
    \delta\chi_{k+1} &= A_k\delta\chi_k + G_k\omega_k, \\
    \delta y_k &= C_k\delta\chi_k + \nu_k,
\end{aligned}
\end{equation}
where $\delta\chi_k := \chi_k - \hat{x}_{k|T-1}$, $\delta y_k := y_k - h(\hat{x}_{k|T-1})$, and
\begin{equation*}
    A_k := \frac{\partial f(\chi,u,\omega)}{\partial \chi} \bigg|_{(\hat{x}_{k|T-1}, u_k, 0)}, \qquad
    G_k := \frac{\partial f(\chi,u,\omega)}{\partial \omega} \bigg|_{(\hat{x}_{k|T-1}, u_k, 0)}, \qquad
    C_k := \frac{\partial h(\chi)}{\partial \chi} \bigg|_{\hat{x}_{k|T-1}},
\end{equation*}for $k=T-N, ~T-N+1, \ldots, ~T-1$. Note that the control input $u$ does not appear in \eqref{eq:dynamics_LTV} since it is a constant in the estimation problem, and therefore disappears with linearization. The approximated linear \ac{mhe} problem at time $T \geq N$ is

\begin{equation}
\begin{aligned}
    \min_{\delta\chi_{T-N},\{\omega_k\}_{k=T-N}^{T-1}} &V\big(\delta\chi_{T-N}, \{\omega_k\}_{k=T-N}^{T-1}, T\big) \\
    \text{s.t. } \eqref{eq:dynamics_LTV},
\end{aligned}
\end{equation}
which is a minimization problem with a quadratic objective function and affine equality constraints. We rewrite this problem as a standard unconstrained \acl{qp} (\acs{qp})

\begin{equation}\label{eq:lmhe_qp}
    \min_{\mathbf{z}} \frac{1}{2}\mathbf{z}^\top \mathbf{H} \mathbf{z} + \mathbf{f}^\top \mathbf{z},
\end{equation}
with the decision variable $\mathbf{z}=(\delta\chi_{T-N}^\top, \{\omega_k^\top\}_{k=T-N}^{T-1})^\top \in \R^{(N+1)n}$. Note that instead of treating the dynamics \eqref{eq:dynamics_LTV} as equality constraints, which can increase the dimension of the problem, this formulation propagates the dynamics from $\chi_{T-N}$ within the estimation window and directly incorporates it into the objective function, producing a smaller but dense unconstrained \ac{qp}. This is often referred to as the \emph{condensing approach} \cite{Kouzoupis2018QP}. For the smoothing \ac{mhe} with arrival cost adjustment, the expressions for the objective function, leaving out the constant terms which do not affect the minimizer, are given by
\begin{equation}\label{eq:lmhe_qp_H_f}
\begin{aligned}
    \mathbf{H} &= \mathbf{P}^{-1} + \mathbf{Q}^{-1} + \mathbf{G}^\top\mathbf{A}^\top\mathbf{C}^\top \mathbf{R}^{-1}\mathbf{C}\mathbf{A}\mathbf{G} - \mathbf{H}_{smooth}, \\
    \mathbf{f} &= \overline{\mathbf{P}} - \mathbf{G}^\top\mathbf{A}^\top\mathbf{C}^\top \mathbf{R}^{-1} \mathbf{\delta y} - \mathbf{f}_{smooth},
\end{aligned}
\end{equation}
where
\begin{gather*}
    \mathbf{P}^{-1} = \begin{pmatrix}
        P_{T-N}^{-1} & \\
        & \mathbf{0}_{Nn \times Nn}
    \end{pmatrix}, \quad
    \mathbf{Q}^{-1} = \begin{pmatrix}
        \mathbf{0}_{n\times n} &&& \\
        & Q_{T-N}^{-1} && \\
        && \ddots & \\
        &&& Q_{T-1}^{-1}
    \end{pmatrix}, \quad
    \mathbf{G} = \begin{pmatrix}
        \mathbf{I}_{n} &&& \\
        & G_{T-N} && \\
        && \ddots & \\
        &&& G_{T-1}
    \end{pmatrix}, \\
    \mathbf{A} = \mathbf{A}_\chi + \mathbf{A}_\omega, \quad
    \mathbf{C} = \begin{pmatrix}
        C_{T-N} &&& \\
        & C_{T-N+1} && \\
        && \ddots & \\
        &&& C_T
    \end{pmatrix}, \quad
    \mathbf{R}^{-1} = \begin{pmatrix}
        R_{T-N}^{-1} &&& \\
        & R_{T-N+1}^{-1} && \\
        && \ddots & \\
        &&& R_T^{-1}
    \end{pmatrix}, \\
    \mathbf{A}_\chi = \left(\begin{array}{c|c}
        \mathbf{I}_{n} & \\
        A_{T-N} & \\
        A_{T-N+1}A_{T-N} & \mathbf{0}_{(N+1)n \times Nn} \\
        \vdots & \\
        \prod\limits_{i=T-1}^{T-N} A_i &
    \end{array}\right), \quad
    \mathbf{A}_\omega = \left(\begin{array}{c|ccccc}
         & \mathbf{0}_{n \times n} &&& \\
         & \mathbf{I}_{n} &&& \\
        \mathbf{0}_{(N+1)n \times n} & A_{T-N+1} & \mathbf{I}_{n} && \\
         & \vdots & \vdots & \ddots & \\
         & \prod\limits_{i=T-1}^{T-N+1} A_i & \prod\limits_{i=T-1}^{T-N+2} A_i & \cdots & \mathbf{I}_{n}
    \end{array}\right), \\
    \overline{\mathbf{P}} = \begin{pmatrix}
        -P_{T-N}^{-1}\hat{x}_{T-N} \\ \mathbf{0}_{Nn \times 1}
    \end{pmatrix}, \quad
    \mathbf{\delta y} = \begin{pmatrix} \delta y_{T-N} \\ \delta y_{T-N+1} \\ \vdots \\ \delta y_T \end{pmatrix},
\end{gather*}
where $\hat{x}_{T-N}:=\hat{x}_{T-N|T-1}$, and $P_{T-N}:=P_{T-N|T-1}$ can be calculated using the backward iteration \eqref{eq:mhe_smoothing_P_backward_iteration}.

The formulae for the adjustment terms are
\begin{equation}\label{eq:lmhe_qp_H_f_adjustment}
\begin{aligned}
    \mathbf{H}_{smooth} &= \mathbf{G}^\top\mathbf{A}_\chi^\top\mathbf{\overline{C}}^\top \mathbf{W}^{-1} \mathbf{\overline{C}}\mathbf{A}_\chi\mathbf{G}, \\
    \mathbf{f}_{smooth} &= - \mathbf{G}^\top\mathbf{A}_\chi^\top\mathbf{\overline{C}}^\top \mathbf{W}^{-1} \mathbf{\delta\overline{y}},
\end{aligned}
\end{equation}
where
\begin{gather*}
     \mathbf{\overline{C}} = \left(\begin{array}{cccc|c}
        C_{T-N} &&& & \\
        & C_{T-N+1} && & \mathbf{0}_{Np \times n} \\
        && \ddots & & \\
        &&& C_{T-1} &
    \end{array}\right), \quad
    \mathbf{W} = \mathbf{\overline{C}}\mathbf{A}_\omega\mathbf{G} \mathbf{\overline{Q}} \mathbf{G}^\top\mathbf{A}_\omega^\top \mathbf{\overline{C}}^\top  + \mathbf{\overline{R}}, \quad
     \mathbf{\delta\overline{y}} = \begin{pmatrix} \delta y_{T-N} \\ \delta y_{T-N+1} \\ \vdots \\ \delta y_{T-1} \end{pmatrix}, \\
    \mathbf{\overline{Q}} = \begin{pmatrix}
        Q_{T-N} &&& \\
        & \ddots && \\
        && Q_{T-1} & \\
        &&& \mathbf{0}_{n\times n}
    \end{pmatrix}, \quad
    \mathbf{\overline{R}} = \begin{pmatrix}
        R_{T-N} &&& \\
        & R_{T-N+1} && \\
        && \ddots & \\
        &&& R_{T-1}
    \end{pmatrix}.
\end{gather*}

Note that $\mathbf{H}_{smooth}$ and $\mathbf{f}_{smooth}$ only affect the upper left $n\times n$ block of $\mathbf{H}$ and first $n$ elements of $\mathbf{f}$. Since $\mathbf{H}$ is symmetric, \eqref{eq:lmhe_qp} is a convex problem if and only if $\mathbf{H}$ is positive semi-definite. This condition is always satisfied for the \emph{filtering} \ac{mhe}, i.e., when $\mathbf{H}_{smooth}=0$ and $\mathbf{f}_{smooth}=0$. For the \emph{smoothing} \ac{mhe}, tuning of the weighting matrices $Q,R,P_0$ must be performed carefully with respect to the horizon $N$, so that $\mathbf{H}$ is positive semi-definite. As a basic guideline for tuning, $P_0$ should be small for large $N$.

There are several methods for solving \eqref{eq:lmhe_qp}. It is a standard \ac{qp}, which is supported by many solvers such as OSQP \cite{osqp}, or Clarabel \cite{Clarabel_2024}. However, these numerical solvers often require multiple iterations to reach the optimum; this is not a problem for modern computers with powerful processors, but for online state estimation on small aerial robotic systems with limited computational power, such as quadrotors, faster methods for solving the optimization problem might be required to ensure real-time performance.

% -----------------------------------

\subsection{\acl{l1ao}}\label{sec:methodology_l1ao}

In \ac{mhe}, at each time step, as the measurements and control inputs get updated, we get a new optimization problem. Standard approaches treat these as separate problems, each of which requires solvers performing multiple iterations to find the optimum. Time-varying methods, on the other hand, treat the optimization problem as time-varying and exploit its temporal change by a prediction term, tracking the optimal solutions with only one iteration for each time step. For \eqref{eq:lmhe_qp}, we consider the objective function to be time-varying, i.e.,
\begin{equation}\label{eq:lmhe_qp_time_varying}
    \min_{\mathbf{z}(t)} V(\mathbf{z}(t),t) = \min_{\mathbf{z}(t)} \frac{1}{2}\mathbf{z}^\top(t) \mathbf{H}(t) \mathbf{z}(t) + \mathbf{f}^\top(t) \mathbf{z}(t).
\end{equation}

\begin{assumption}\label{ass:positive_definite_H}
$\mathbf{H}(t)$ is uniformly positive definite, i.e., $\exists m_f>0$ such that $\mathbf{H}(t)\succeq m_f\mathbf{I} ~\forall t\geq0$.
\end{assumption}

Consider the \ac{qp} \eqref{eq:lmhe_qp_time_varying} and suppose the weighting matrices $Q,R,P_0$ are properly tuned so that \Cref{ass:positive_definite_H} holds. Consider a baseline optimizer given by the continuous-time Newton's method
\begin{equation}\label{eq:l1ao_zb}
    \dot{\mathbf{z}}_b(t) = -\nabla_{\mathbf{z}\mathbf{z}}^{-1} V(\mathbf{z}(t),t) \nabla_{\mathbf{z}} V(\mathbf{z}(t),t)
    = -\mathbf{z}(t) - \mathbf{H}^{-1}(t)\mathbf{f}(t).
\end{equation}

Note that this baseline optimizer does not include any temporal prediction, hence still treats \ac{mhe} as a sequence of independent optimization problems. The \ac{l1ao} \cite{kim2025L} augments this baseline dynamics with a temporal prediction that can be \emph{uncertain}, and an adaptive term that compensates for the gradient prediction inaccuracy while preserving fast convergence. The components of \ac{l1ao} \cite{kim2025L}, namely the gradient predictor, the piece-wise constant adaptation law, and the low-pass filter, adopted for the \ac{qp} \eqref{eq:lmhe_qp_time_varying}, are given by
\begin{subequations}\label{eq:lmhe_qp_with_l1ao}
\begin{align}
    \dot{\hat{\nabla}}_{\mathbf{z}}V(\mathbf{z}(t),t) &= \mathbf{A}_s\Tilde{\nabla}_{\mathbf{z}}V(\mathbf{z}(t),t) + \hat{\nabla}_{\mathbf{z}t}V(\mathbf{z}(t),t) + \mathbf{H}(t)\dot{\mathbf{z}}(t) + \mathbf{h}(t), \qquad \hat{\nabla}_{\mathbf{z}}V(\mathbf{z}(0),0) = \nabla_{\mathbf{z}}V(\mathbf{z}(0),0), \\
    \mathbf{h}(iT_s) &= \Big( \mathbf{A}_s^{-1}\big(\mathbf{I} - e^{\mathbf{A}_sT_s}\big) \Big)^{-1} e^{\mathbf{A}_sT_s} \Tilde{\nabla}_{\mathbf{z}}V(\mathbf{z}(iT_s),iT_s), \qquad \forall i\in\N, ~\mathbf{h}(0)=0, \\
    \hat{\boldsymbol{\sigma}}(iT_s) &= \mathbf{H}^{-1}(iT_s) \mathbf{h}(iT_s), \qquad \forall i\in\N, ~\hat{\boldsymbol{\sigma}}(0)=0, \\
    \dot{\mathbf{z}}_a(s) &= C(s)(-\hat{\boldsymbol{\sigma}}(s)),
\end{align}
\end{subequations}
where $\mathbf{A}_s \in \R^{(N+1)n \times (N+1)n}$ is a diagonal Hurwitz matrix, $T_s$ is the sampling period, $\Tilde{\nabla}_{\mathbf{z}}V := \hat{\nabla}_{\mathbf{z}}V - \nabla_{\mathbf{z}}V$ is the prediction error, $\dot{\mathbf{z}} = \dot{\mathbf{z}}_b + \dot{\mathbf{z}}_a$, $C(s)=\frac{\omega_c}{s+\omega_c}$ is a first-order low-pass filter with cutoff frequency $\omega_c>0$, and $\hat{\nabla}_{\mathbf{z}t}V(\mathbf{z}(t),t)$ is the uncertain temporal prediction of the gradient dynamics. In our \ac{mhe} setting, the temporal prediction is constructed from finite differences as
\begin{equation}\label{eq:grad_zt_finite_differences}
    \hat{\nabla}_{\mathbf{z}t} V(\mathbf{z}(t),t) = \frac{\mathbf{H}(t+T_s) - \mathbf{H}(t)}{T_s} \mathbf{z}(t) + \frac{\mathbf{f}(t+T_s) - \mathbf{f}(t)}{T_s}.
\end{equation} This prediction captures how the optimizer changes over time as new measurements enter the \ac{mhe} window. However, in online implementation, this term can be noisy under the effect of measurement noise, or highly inaccurate due to imperfect prior knowledge of the system or operating condition. In this case, the \ac{l1ao} augmentation introduces temporal prediction into the \ac{mhe} problem, while preventing the performance of the solver from degrading due to prediction inaccuracy.

The behavior of \ac{l1ao} is governed by the design parameters $\mathbf{A}_s$ and $\omega_c$. The matrix $\mathbf{A}_s$ is chosen to ensure stable and sufficiently fast decay of the prediction error $\Tilde{\nabla}_{\mathbf{z}}V$. The cutoff frequency $\omega_c$ trades off adaptation speed and robustness. In our implementation, these parameters are tuned empirically to balance convergence speed, robustness in a wide range of operating conditions, and numerical stability of the \ac{mhe} solver.

% -----------------------------------

\section{Simulations}\label{sec:simulation}

In this section, we perform simulations of different scenarios on the quadrotor model in \Cref{sec:preliminaries_quadrotor} to demonstrate the performance of \ac{mhe} with different choices of solvers and the \ac{ekf}. First, we investigate the estimators in uncertainty-free conditions as a baseline for comparison. Next, we initialize the estimators with bad initial guesses $\overline{x}_0$ to demonstrate their convergence characteristics. Finally, we introduce a process noise with multiple frequency components into the quadrotor, showing the advantages of the moving-horizon approach in comparison with the one-step propagation of the \ac{ekf}.

The simulation parameters are $m=1~\si{kg}, ~g=9.81~\si{m\cdot s^{-2}}, ~J=\diag(5\cdot10^{-3},5\cdot10^{-3},9\cdot10^{-3})~\si{kg\cdot m^2}$, and all measurement channels are subject to zero-mean, normally distributed measurement noise with a standard deviation of $10^{-1}$, i.e., $v \sim \mathcal{N}(0,10^{-2})$. The \ac{mhe} is of smoothing type, with arrival cost computation using \ac{ekf}, and is implemented in the standard \ac{qp} form derived in \Cref{sec:methodology_mhe_qp}. To evaluate the performance of our proposed method, we will compare the \ac{mhe} using the time-varying method described in \Cref{sec:methodology_l1ao} with a standard \ac{mhe} implemented in CVXPY \cite{diamond2016cvxpy} using the solver OSQP \cite{osqp}.
The simulation uses Euler integration with a sampling period of $T_s=10^{-2}$ second. All simulation scenarios run in Python on a laptop with an AMD Ryzen 7 4800U processor. To evaluate the overall estimation accuracy of each method, we calculate their \ac{rmse} over all time steps $k=0,1,2,\ldots, T$ that is given by
\begin{equation}\label{eq:rmse_formula}
    \|x - \hat{x}\|_{RMS} = \sqrt{\frac{1}{T+1} \sum\limits_{k=0}^T \|x_k - \hat{x}_k\|^2}.
\end{equation}

% -----------------------------------

\subsection{Performance in uncertainty-free conditions}\label{sec:simulation_baseline}

First, we compare the baseline performance of the \ac{ekf}, the standard \ac{mhe} using the solver OSQP, and the \ac{l1ao}-based \ac{mhe} in an uncertainty-free condition, i.e., with a perfect initial guess and no process noise. The weighting matrices are chosen as
\begin{equation*}
    P_0=10^{-2}\mathbf{I}_{12}, \qquad Q=10^{-2}\mathbf{I}_{12}, \qquad R=10^{-2}\mathbf{I}_{6}.
\end{equation*}
The \ac{mhe} horizon length is $N=10$, and the parameters of \ac{l1ao} are
\begin{equation*}
    \mathbf{A}_s=-10^2\mathbf{I}_{12(N+1)}, \qquad \omega_c=150.
\end{equation*}

The quadrotor uses a linear quadratic regulator (LQR) controller to track an aggressive circular trajectory with $\|q\|_\infty \leq 3~\si{m\cdot s^{-1}}$, $\|\Theta\|_\infty \leq 85^\circ$, and $\|\Omega\|_\infty \leq 20~\si{rad\cdot s^{-1}}$.

\begin{figure}[hbt!]
    \centering
    \includegraphics[width=0.8\linewidth]{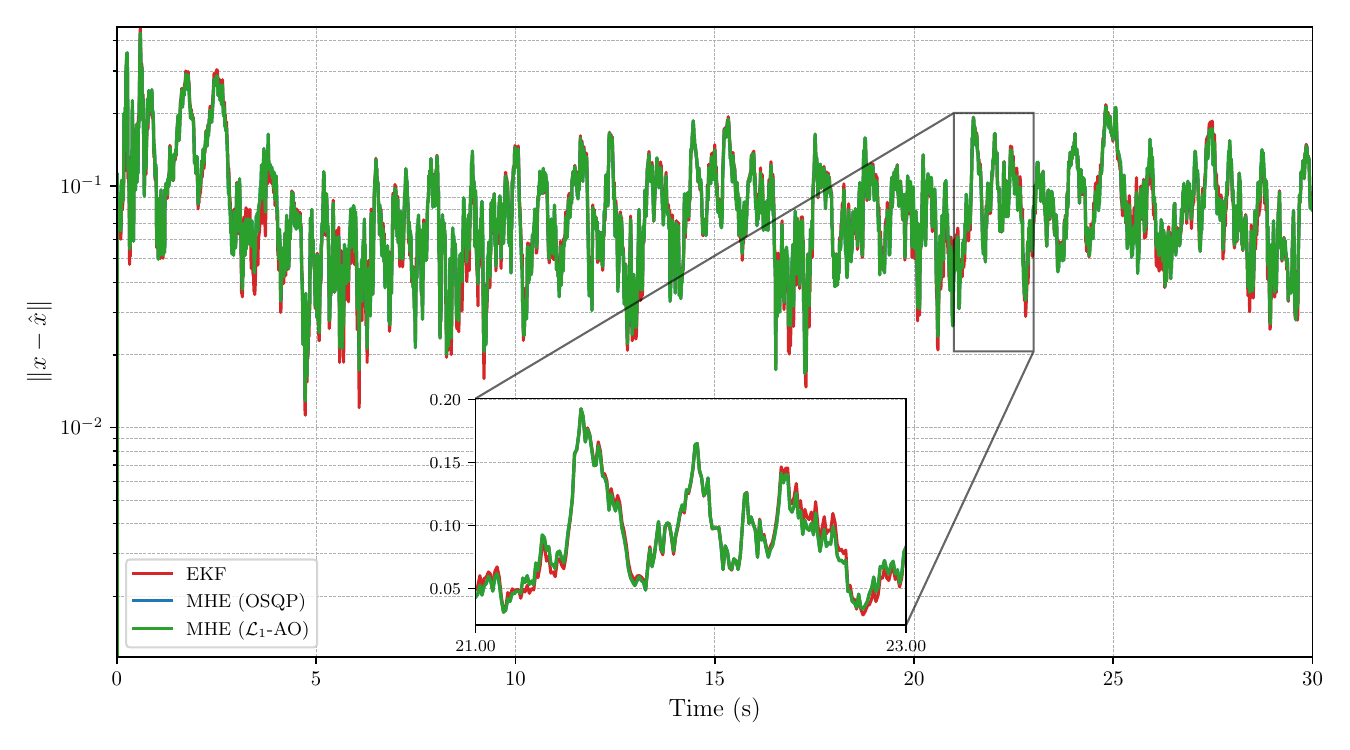}
    \caption{Estimation errors evolution in an uncertainty-free scenario}
    \label{fig:scenario0_errors}
\end{figure}

\textbf{Results:} The simulation results are illustrated in \Cref{fig:scenario0_errors}. As shown in \Cref{fig:scenario0_errors}, all estimators give nearly equivalent performances. During a $30$-second long simulation with $3001$ data points, the estimation \ac{rmse} of the \ac{ekf} is $0.0989$, while that of the standard \ac{mhe} and the \ac{l1ao}-based \ac{mhe} are both $0.0985$. Although it is natural to expect \ac{mhe} to provide a superior estimation performance compared to the \ac{ekf}, this result shows that in scenarios where we have perfect knowledge of every parameter of the system and operating condition, the \ac{ekf} and \ac{mhe} can give equivalent performances, and the choice of solvers does not improve or degrade the performance of \ac{mhe}.

% -----------------------------------

\subsection{Estimation error convergence given poor initial guesses}\label{sec:simulation_bad_initial guess}

In practice, the estimators are often initialized with an initial guess $\overline{x}_0$ that differs from the true system state $x_0$. If the controller uses the estimate, slow convergence of the estimation error might lead to undesirable transient characteristics; hence, we would like the estimate to converge to the true state as quickly as possible, with minimal oscillations. 

We run $100$ simulations with randomly generated poor initial guesses, and the estimator initialization error is normalized to $10$ for consistency, i.e., $\|x_0-\overline{x}_0\|=10$. The quadrotor is set to hover to demonstrate only the convergence properties of each estimator. The weighting matrices are chosen as
\begin{equation*}
    P_0=10^{-2}\mathbf{I}_{12}, \qquad Q=10^{-2}\mathbf{I}_{12}, \qquad R=10^{-2}\mathbf{I}_{6}.
\end{equation*}
The \ac{mhe} horizon length is $N=10$, and the parameters of \ac{l1ao} are
\begin{equation*}
    \mathbf{A}_s=-10^{-1}\mathbf{I}_{12(N+1)}, \qquad \omega_c=150.
\end{equation*}

\begin{figure}[hbt!]
    \centering
    \includegraphics[width=0.8\linewidth]{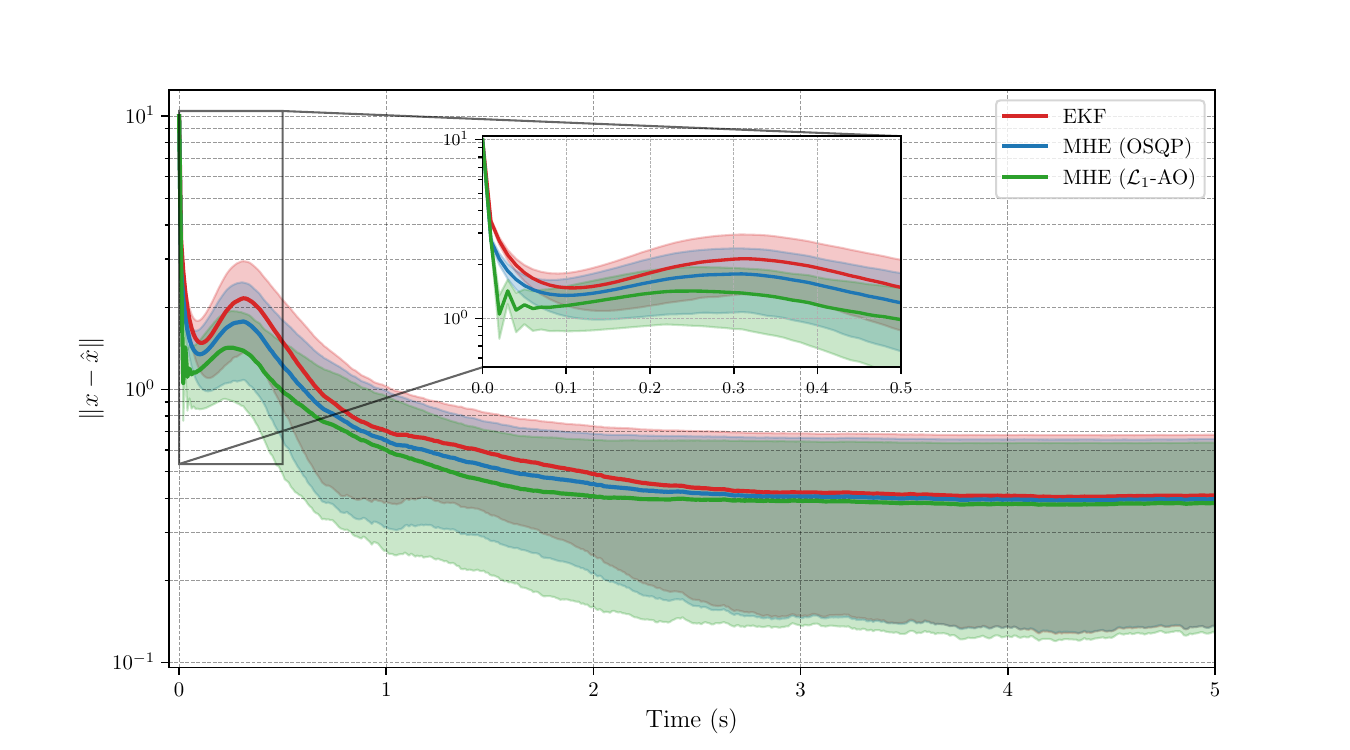}
    \caption{Mean and standard deviation of estimation errors over $100$ simulations with poor initial guesses}
    \label{fig:scenario1_errors}
\end{figure}

\textbf{Results:} The simulation results for $100$ runs are illustrated in \Cref{fig:scenario1_errors}. Each line represents the mean of estimation error of each estimator over all runs, and the range within one standard deviation from the mean is shaded using the same color. Both types of smoothing \ac{mhe} provide faster estimation error convergence to zero compared to the \ac{ekf}. Among the estimators, the \ac{l1ao}-based \ac{mhe} gives the fastest error convergence, while the \ac{ekf} gives the slowest. \Cref{fig:scenario1_errors} also shows that all estimators exhibit a common transient behavior at $0.2$ second, where the error means increase before converging to their respective steady-state values, with the \ac{l1ao}-based \ac{mhe} always having the lowest mean. Over $100$ $5$-second long simulations, the mean \ac{rmse} for the \ac{ekf}, the standard \ac{mhe}, and the \ac{l1ao}-based \ac{mhe} are $0.8604$, $0.7633$, and $0.6432$, respectively. The proposed \ac{mhe} with \ac{l1ao} provides $25.2\%$ and $15.7\%$ improvement in \ac{rmse} compared to the \ac{ekf} and the standard \ac{mhe}, respectively. This shows that given a poor initial guess, solving the \ac{mhe} problem with \ac{l1ao} can provide faster convergence and lower overall estimation \ac{rmse}.

% -----------------------------------

\subsection{Performance with unknown process noise}\label{sec:simulation_unknown_process_noise}

In real-world quadrotors, multiple factors can contribute to process noise, including modeling errors, unmodeled dynamics, and external disturbances. In this scenario, the quadrotor tracks the circular trajectory described in \Cref{sec:simulation_baseline}, while being injected with the following process noise $w(t)$ with an amplitude of $1.0$ in the linear acceleration channels
\begin{equation*}
    w(t) = \begin{pmatrix}
        \mathbf{0}_{3 \times 1} \\
        0.33\sin(6.57t+1.88) + 0.33\sin(6.34t+3.87) + 0.33\sin(8.30t+4.33) \\
        0.33\sin(3.69t+2.66) + 0.33\sin(7.30t+2.41) + 0.33\sin(1.83t+2.44) \\
        0.33\sin(1.89t+0.18) + 0.33\sin(5.84t+6.27) + 0.33\sin(7.30t+0.85) \\
        \mathbf{0}_{3 \times 1} \\
        0.17\sin(1.70t+0.78) + 0.17\sin(8.91t+6.16) + 0.17\sin(2.95t+4.53) \\
        0.17\sin(7.96t+4.21) + 0.17\sin(7.98t+4.31) + 0.17\sin(8.35t+3.30) \\
        0.17\sin(8.74t+4.07) + 0.17\sin(1.59t+4.09) + 0.17\sin(5.82t+1.95)
    \end{pmatrix},
\end{equation*}  and $0.5$ in the angular acceleration channels. The weighting matrices are chosen as
\begin{equation*}
    P_0=10^{-2}\mathbf{I}_{12}, \qquad R=10^{-1}\mathbf{I}_{6},
\end{equation*}
and multiple values of $Q$ and horizon length $N$ are chosen to demonstrate their effects on the performance of \ac{mhe}. The parameters of \ac{l1ao} are
\begin{equation*}
    \mathbf{A}_s=-10^2\mathbf{I}_{12(N+1)}, \qquad \omega_c=150.
\end{equation*}

\begin{figure}[hbt!]
    \centering
    \includegraphics[width=0.8\linewidth]{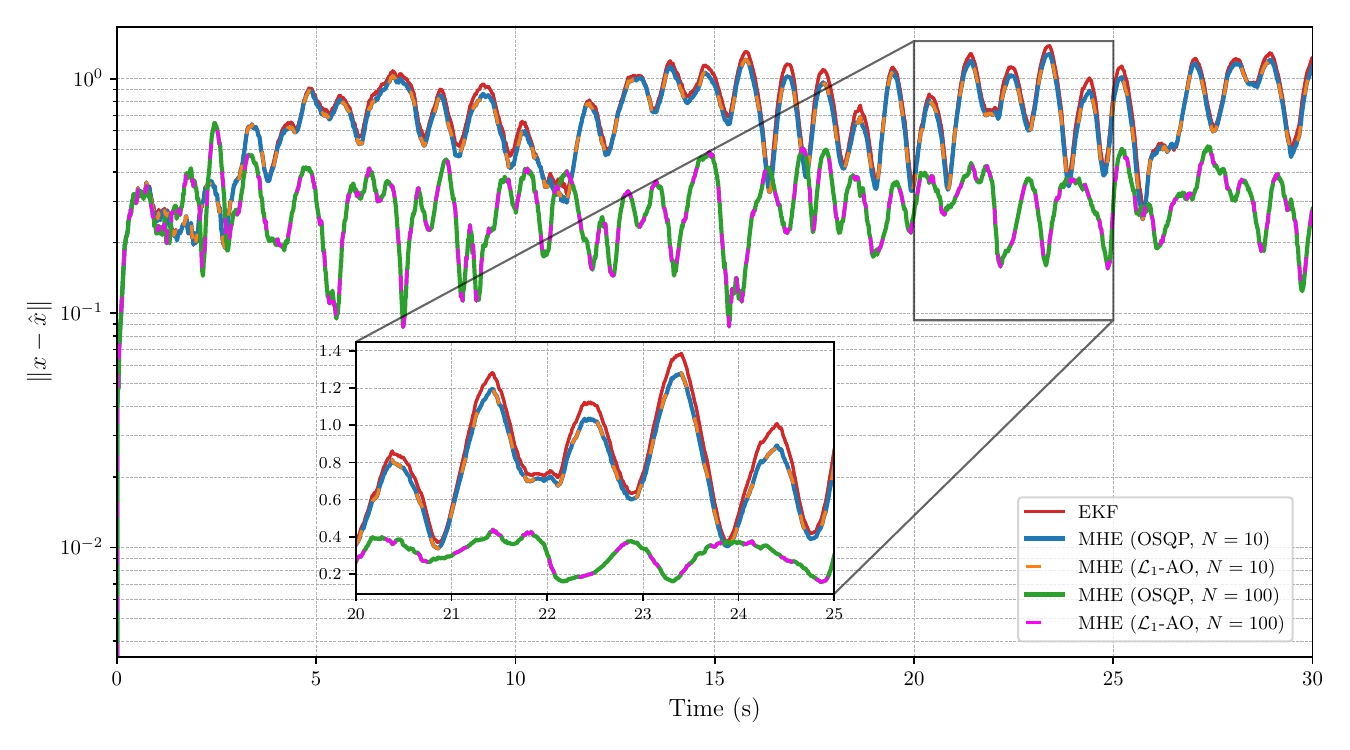}
    \caption{Estimation errors evolution given unknown process noise ($Q=10^{-2}\mathbf{I}_{12}$)}
    \label{fig:scenario2_errors}
\end{figure}

\begin{table*}[hbt!]
	\centering
	\small
	\vspace{-0.2cm}
	\captionsetup{font=small}
	\caption{Comparison of \ac{rmse} for different values of $Q$ and \ac{mhe} horizon length $N$ with unknown process noise, during a $30$-second long simulation ($3001$ data points). Note that the \ac{rmse}s, calculated using \eqref{eq:rmse_formula}, are unitless.}
    \label{tab:scenario2_rmse_by_Q_N}
    
\begin{tabular}{c|c|cccc}
\toprule
\multirow{2}{*}{$Q$} 
& \multirow{2}{*}{EKF} 
& \multicolumn{4}{c}{\ac{mhe} (\ac{l1ao})} \\
\cline{3-6}
& & $N=10$ & $N=20$ & $N=50$ & $N=100$ \\
\midrule
$10^{-2}\mathbf{I}_{12}$ 
& 0.7769 & 0.7269 & 0.5974 & 0.3208 & 0.3109 \\ \hline

$\diag(10^{-2}\mathbf{I}_3,~10^{-1}\mathbf{I}_3,~10^{-2}\mathbf{I}_3,~10^{-1}\mathbf{I}_3)$ 
& 0.3427 & 0.3190 & 0.2953 & 0.2927 & 0.2770 \\ \hline

$\diag(10^{-2}\mathbf{I}_3,~\mathbf{I}_3,~10^{-2}\mathbf{I}_3,~\mathbf{I}_3)$ 
& 0.2834 & 0.2793 & 0.2768 & 0.2368 & 0.2395 \\ 
\bottomrule
\end{tabular}
\end{table*}

\begin{figure}[hbt!]
    \centering
    \includegraphics[width=\linewidth]{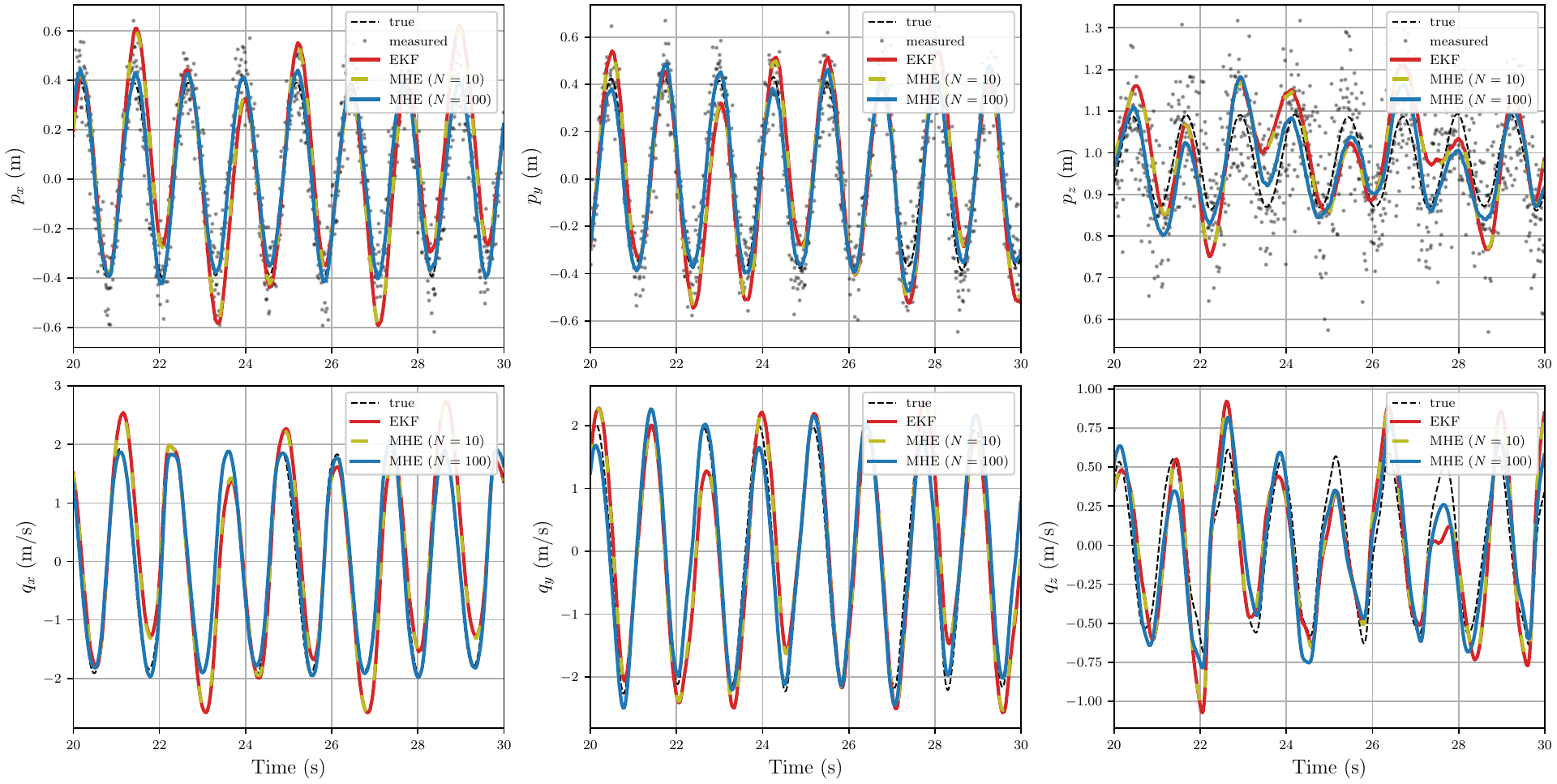}
    \caption{Comparison between estimates of the linear position $p$ and velocity $q$ of the \ac{ekf} and the \ac{l1ao}-based \ac{mhe} given unknown process noise ($Q=10^{-2}\mathbf{I}_{12}$)}
    \label{fig:scenario2_states_linear}
\end{figure}

\begin{figure}[hbt!]
    \centering
    \includegraphics[width=\linewidth]{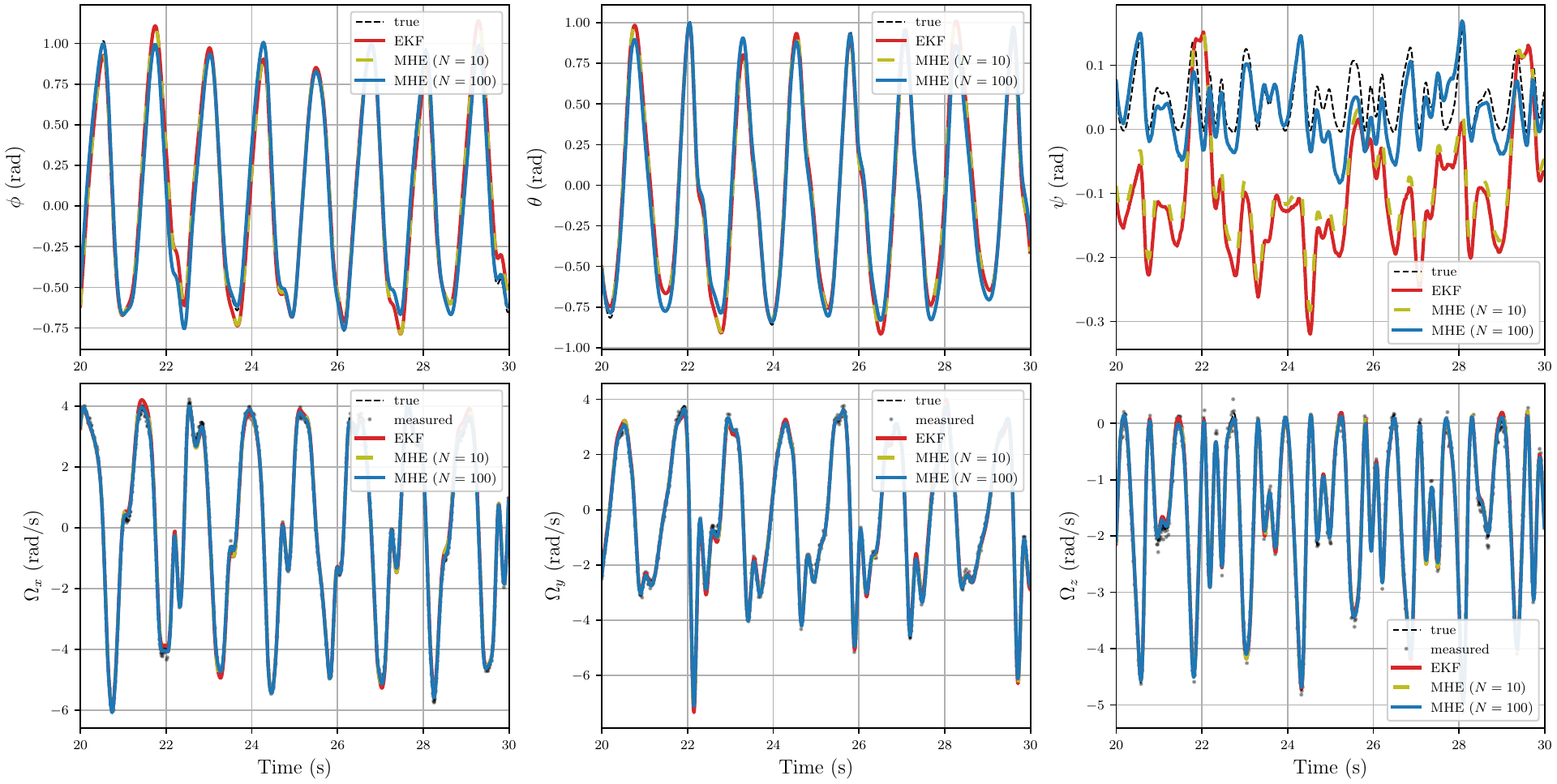}
    \caption{Comparison between estimates of the Euler angles $\Theta$ and angular velocity $\Omega$ of the \ac{ekf} and the \ac{l1ao}-based \ac{mhe} given unknown process noise ($Q=10^{-2}\mathbf{I}_{12}$)}
    \label{fig:scenario2_states_angular}
\end{figure}

\textbf{Results:} \Cref{fig:scenario2_errors} illustrates the estimation errors over time for the case where $Q=10^{-2}\mathbf{I}_{12}$. Due to the optimization-based structure, both types of \ac{mhe} with different solvers effectively estimate the unknown process noise and offer significantly better performance compared to the \ac{ekf}. In this case, both types of \ac{mhe} give the exact same performance. \Cref{tab:scenario2_rmse_by_Q_N} shows the estimation \ac{rmse} for different tuning values of $Q$ and horizon length $N$. As a general observation, as we increase $Q$, all estimators rely more on the measurements and therefore give better results. However, \ac{mhe} consistently gives lower \ac{rmse} than the \ac{ekf}, and the improvement can be as high as $60\%$ in the case where $Q=10^{-2}\mathbf{I}_{12}$ and $N=100$. \Cref{fig:scenario2_states_linear} illustrates the estimates of each component of the linear position $p$ and the linear velocity $q$ of the \ac{ekf}, the \ac{l1ao}-based \ac{mhe} with a short horizon ($N=10$), and that with a long horizon ($N=100$). While the \ac{ekf} and the short-horizon \ac{mhe} give comparable estimates with frequent overshoots, the long-horizon \ac{mhe} generates more accurate estimates that are closer to the true system states. \Cref{fig:scenario2_states_angular} illustrates the estimates of each component of the Euler angles $\Theta$ and the angular velocity $\Omega$. An immediate observation in \Cref{fig:scenario2_states_angular} is that the yaw angle $\psi$ estimates of the \ac{ekf} and the short-horizon \ac{mhe} are highly inaccurate, although the short-horizon \ac{mhe} estimate is slightly closer to the true yaw angle. Compared to the \ac{ekf} and the short-horizon \ac{mhe}, the long-horizon \ac{mhe} generates highly accurate estimates of all of the angular components, demonstrating its effectiveness.

We see that the performance of \ac{mhe} can be greatly improved with longer horizon lengths. However, this improvement comes at the cost of high computation, which is shown in \Cref{tab:scenario2_computation_time}. The computation time of OSQP more than doubles that of \ac{l1ao}, and the difference grows as the horizon length increases. This is due to the fact that at each time step, OSQP performs multiple iterations to reach the optimum, while \ac{l1ao} with the baseline continuous-time Newton's method only solves a linear system, offering the benefits of less computation workload.
This implies that the \ac{l1ao}-based \ac{mhe} is more efficient and suitable for real-time applications.

\begin{table*}[hbt!]
	\centering
	\small
	\vspace{-0.2cm}
	\captionsetup{font=small}
	\caption{Mean computation time per time step (in milliseconds) for \ac{mhe} with different horizon lengths $N$.}
    \label{tab:scenario2_computation_time}
    
\begin{tabular}{c|c|c}
\toprule
$N$ & OSQP & \ac{l1ao} \\ 
\midrule
$10$ & 11.16 & 4.34 \\ \hline
$20$ & 38.53 & 18.67 \\ \hline
$50$ & 233.22 & 88.25 \\ \hline
$100$ & 1373.45 & 405.30 \\ 
\bottomrule
\end{tabular}

\end{table*}

% -----------------------------------

\section{Conclusion}\label{sec:conclusion}

In this study, we formulated a linear smoothing \acl{mhe} \acl{qp} and proposed a time-varying solver consisting of the Newton's method and the \acl{l1ao} to efficiently solve the estimator optimization problem. The smoothing \ac{mhe} problem is formulated in the standard form of a \acl{qp}, which allows for flexible choices of solving methods. Through numerical simulations on a quadrotor platform, we demonstrated that the proposed time-varying solver accelerates convergence of the estimation error with poor initial guesses, and runs more than two times faster than time-invariant solvers that rely on iterative methods.

Future work includes approximating the \ac{mhe} arrival cost using sampling-based methods, such as the Unscented Kalman Filter. This could further enhance estimation accuracy while enabling shorter horizons, making \ac{mhe} more practical for real-time deployment. We also aim to explore constrained state estimation, where the benefits of \ac{l1ao} and other time-varying interior-point optimization methods can be leveraged further.

% -----------------------------------

\section*{Acknowledgments}
This work is supported by the Air Force Office of Scientific Research Grant FA9550-21-1-0411, the National Aeronautics and Space Administration under Grant 80NSSC22M0070, and the National Science Foundation (NSF) under Grants CMMI 2135925, IIS 2331878, and  CMMI 24-31216. We also acknowledge the Vingroup Science and Technology Scholarship Program as the scholarship provider for T. Q. Nguyen's Master's program. 

\bibliography{references.bib}

\end{document}